*Article*

# GRUvader: Sentiment-Informed Stock Market Prediction

Akhila Mamillapalli [1], Bayode Ogunleye [1,*], Sonia Timoteo Inacio [1] and Olamilekan Shobayo [2]

1. School of Architecture, Technology and Engineering, University of Brighton, Brighton BN2 4GJ, UK; s.timoteoinacio@brighton.ac.uk (S.T.I.)
2. School of Computing and Digital Technologies, Sheffield Hallam University, Sheffield S1 2NU, UK; o.shobayo@shu.ac.uk
* Correspondence: b.ogunleye@brighton.ac.uk

**Abstract:** Stock price prediction is challenging due to global economic instability, high volatility, and the complexity of financial markets. Hence, this study compared several machine learning algorithms for stock market prediction and further examined the influence of a sentiment analysis indicator on the prediction of stock prices. Our results were two-fold. Firstly, we used a lexicon-based sentiment analysis approach to identify sentiment features, thus evidencing the correlation between the sentiment indicator and stock price movement. Secondly, we proposed the use of GRUvader, an optimal gated recurrent unit network, for stock market prediction. Our findings suggest that stand-alone models struggled compared with AI-enhanced models. Thus, our paper makes further recommendations on latter systems.

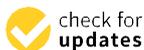

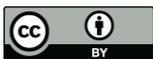

## 1. Introduction

The general public are increasingly interested in stock market prediction and, more specifically, the traders, investors, and decision-makers [1]. This is due to their interest in making informed and strategic decisions on their financial stock business, investments, and intentions to invest. The stock market influences the economic growth of every nation, enabling the buying and selling of investments, which supports businesses and boosts the economy [2]. Therefore, accurate stock market prediction is vital to understand the market and mitigate investment loss. Recent advancements in technology and machine learning (ML) techniques have improved stock market prediction [1]. However, due to the unpredictive nature of stock markets, the influence of external factors, and the high complexity and volatility (short-term fluctuations and long-term trends), enabling a more stable and accurate prediction of stock markets is challenging. Traditional models like autoregressive integrated moving average (ARIMA) have shown effectiveness in short-term forecasting [3]. ARIMA is useful for capturing trends and seasonality [4]; however, it is limited when capturing complex and non-linear relationships. Furthermore, theories like the efficient market hypothesis (EMF) and random walk theory suggest that historical data are insufficient when predicting stock prices because the prices randomly change and are influenced by external factors such as financial news [5]. Improvements in AI have led to the application of neutral-network-based prediction models such as recurrent neural networks (RNNs) and convolutional neural networks (CNNs), which are effective with time series data but face challenges with unstructured (text) data [6,7].

Recent studies have emphasized the importance of incorporating textual data because investor sentiment or external events (for example, Brexit and COVID-19) significantly impact stock price movement [8]. Previous studies [9–11] have explored various ML models for stock prediction, incorporating the text data sentiment. However, these models exhibited low performance for predicting stock prices. Further to this, our literature synthesis (presented in Section 2) indicates that only a few methodologies were robust to exogenous variables from sources such as financial news, company news, market sentiment, public opinion, and investor sentiment. In addition, the literature review findings suggest that despite extensive studies on the application of ML algorithms in this context, optimal stock prediction models are limited. For example, [11] used a regression model to predict stock movement based on a sentiment indicator, but did not clarify which stocks were analyzed and no information was provided about exogenous features towards the behavior of the target variable. Additionally, there was no comparison conducted against other models, nor was optimality achieved.

To this end, our study aims were to develop a stock price prediction model and determine the influence of public sentiment. To achieve this, the study formulated this problem as a regression task. In doing so, we utilized the stock dataset collected by Yukhymenko [11] from 30 September 2021 to 29 September 2022 and compared our approach with several state-of-the-art (SOTA) stock price prediction models, including traditional methods like ARIMA. Therefore, our results were two-fold. The first results were those from the experiment conducted to compare ML algorithms using the stock price dataset only. The second results were those from the experiment that incorporated public sentiment. The latter experiment helped to provide information on the behavior of exogenous features towards stock price. Based on our comprehensive experimentation, we identified that $GRU_{vader}$ was the best-performing stock prediction model. This paper confirmed that $GRU_{vader}$ has a simple architecture (which is fast, computationally inexpensive, and adaptable to noisy data) and can serve as a general-purpose stock price prediction model. This implied that $GRU_{vader}$ could detect sentiment from different stock-related text datasets and, most importantly, would be efficient at handling sequential data for short- and long-term stock price forecasting. The results of this study not only offer insights into the comparative effectiveness of different ML techniques, but also provide valuable guidance for the selection of optimal models based on specific data characteristics. Specially, the findings identified the model with superior accuracy, providing useful insights for stock market stakeholders regarding the key drivers of stock prices. The remainder of this paper is organized into four primary sections. Section 2 focuses on the literature review, Section 3 outlines the methodology, the following sections present the results, and the final section provides conclusions and recommendations for future research.

## 2. Related Works

Stock price movement can be modelled in several ways. There are two popular concepts to this. The first is to model a classification task where the direction of the movement of stock prices can be predicted as "*up*" or "*down*". The second is to predict the stock prices by formulating the problem as a regression task. The former approach is suitable for decision-making processes, especially when beneficiaries are only interested in stock movement indicators. A popular ML algorithm for this approach is logistic regression [12]. The latter approach is more beneficial because it provides an understanding of financial gains or losses in the form of estimates and, therefore, it is more popular. Predicting stock prices provides valuable insights into financial market trends, guides investment decisions, and supports the development of effective financial strategies. Studies like [13–18] have highlighted the reliance of investors and portfolio managers on stock price prediction to make informed investment decisions. Accurate stock price prediction allows investors to proactively adopt their portfolios, capitalize on emerging opportunities, and strategically manage risks, ultimately resulting in more robust investment outcomes. Conventional statistical learning algorithms such as linear regression have been used to model stock price movement [19]. Further to this, other techniques like the moving average [20] and exponential smoothing [21] have also been applied. Moreover, time series models such as generalized autoregressive conditional heteroskedasticity [22,23] and ARIMA [24] have been extensively used. However, these models have been criticized for their limitations regarding model complex patterns within the stock price dataset [25].

More recently, advancements in technology have led to the development and application of ML algorithms for stock price prediction. These ML algorithms—ranging from traditional ML algorithms (K-nearest neighbor [26], support vector regression [27], decision tree [28], random forest [29], and XGboost [30]) to deep learning algorithms (RNN [31], LSTM [32], Bi-LSTM [33], GRU [34], Bi-GRU [35], and GAN [36])—have been used to model stock price movement. In a comparative experimental study, the authors of [37] showed that feed-forward neural networks yielded better results compared with ARIMA and ARIMAX. Furthermore, the long short-term memory (LSTM) network introduced by [38] is a type of RNN designed to effectively manage both long-term and short-term dependencies. Several studies have demonstrated the effectiveness of LSTM in stock prediction. For example, the authors of [39] showed that attention-based LSTM outperformed conventional LSTMs in predicting stock market movements. Study [40] showed that LSTM could be used to enhance predictive performance by leveraging both past and future sequences in a time series. Similarly, the authors of [41] combined Bi-LSTM with a CNN, showing a 9% improvement in prediction performance compared with single pipeline models. The authors of [42] proposed a simple GAN model with LSTM as a generator and a CNN as the discriminator, incorporating 13 technical indicators, and demonstrated that the generative adversarial network (GAN) outperformed other models like LSTM, artificial neural network (ANN), support vector machine (SVM), and ARIMA for stock price prediction. Ref. [43] used a GAN with an LSTM generator and a multiple perceptron (MLP) discriminator, outperforming traditional a model based on RMSE and MAE. The authors of [44] proposed a gated recurrent unit (GRU) with LSTM and a dropout layer to model stock prices based on historical data from S&P 500. Their results demonstrated better performance compared with the RNN, CNN, and other ensembled techniques. Further to this, study [45] indicated that integrating news data with stock data and feeding them into a GRU model led to enhanced prediction results, surpassing LSTM models in accuracy. The authors of [46] combined a GRU and LASSO dimensionality reduction, achieving improved accuracy for stock price prediction using data from the Shanghai Composite Index.

More recently, the authors of [47,48] emphasized the influence of financial news and investors' words on stock price movement. They further stated that stock price prediction was more challenging due to these constraints and, as such, the researchers considered modelling stock price movement by integrating sentiment indicators and financial stock price datasets. A sentiment indicator can be obtained by performing a sentiment analysis of financial news or the words of stakeholders/investors. A sentiment analysis is the process of classifying emotions or opinions, expressing natural language into categories such as positive, negative, and neutral [49–51]. The authors of [40,47,48] asserted that the sentiment analysis of financial news or stakeholder/investor comments is crucial for the forecasting of stock trends. The importance of sentiment analyses for stock prediction has been emphasized in previous studies. For example, the authors of [8] incorporated investor sentiment to predict the CS1300 index using LSTM. The LSTM model outperformed the traditional time series models for financial data. Moreover, the authors of [52,53] demonstrated how incorporating a lexicon-based sentimental analysis of financial news with historical stock data could effectively predict market trends for indices like DIJA and S&P 500. Similarly, [54] found that combining a sentiment analysis with stock data to model stock, particularly a CNN, produced good results. The authors of [55] showed the efficiency of using support vector regression (with multiple kernel learning) and SentiWordNet to model stock price movement. Further to this, study [11] used a GAN to model stock prices; however, their study failed to provide details on the influence of the sentiment indicator utilized. To conclude, it is worth stating that the majority of these studies ignored the use of hyperparameter tuning. In addition, most existing studies did not provide explicit or comprehensive details of the sentiment score implemented. As such, it is difficult to assess the reliability and efficiency of the models.

## 3. Methodology

This paper proposed the use of $GRU_{vader}$ for stock price prediction. In doing so, we compared $GRU_{vader}$ with a traditional time series model (namely, ARIMA) and state-of-the-art ML models. The selection of the ML algorithms was based on their performances shown in the stock market prediction literature, as reviewed in the previous section. In subsequent sections, this study presents details of the datasets and our approach.

*3.1. Datasets*

This study used two datasets to model stock market prediction. The first was the stock price dataset and the second was the text data. Below, we provide an overview of the datasets.

3.1.1. Stock Price Dataset

The stock price dataset was collected by Yukhymenko [11]. This dataset has been extensively used in the literature [9,56] to model stock prices. The dataset contains the daily information of 25 companies from 30 September 2021 to 29 September 2022. The dataset includes 6300 observations over eight features (namely, date, open, high, low, close, adjusted close, volume, and stock name). The data were collected only on weekdays (Monday to Friday), and no missing values were found. The companies represented in the dataset were Tesla (TSLA), Microsoft (MSFT), Amazon (AMZN), Google (GOOG), Apple (AAPL), Netflix (NFLX), Taiwan Semiconductor Manufacturing Company Limited (TSM), Coca Cola (KO), Ford (F), Procter & Gamble Company (PG), Costco (COST), Disney (DIS), Verizon (VZ), Salesforce (CRM), Intel (INTC), Boeing (BA), Blackstone (BX), Northrop Grumman Corporation (NOC), PayPal (PYPL), Enphase Energy (ENPH), NIO In (NIO), ZS, and XPeng Inc (ZPEV). Each company contributed approximately 252 records. Figure 1 presents a view of the dataset.

|  | Date | Open | High | Low | Close | Adj Close | Volume | Stock Name |
|---|---|---|---|---|---|---|---|---|
| 0 | 2021-09-30 | 260.333344 | 263.043335 | 258.333344 | 258.493347 | 258.493347 | 53868000 | TSLA |
| 1 | 2021-10-01 | 259.466675 | 260.260010 | 254.529999 | 258.406677 | 258.406677 | 51094200 | TSLA |
| 2 | 2021-10-04 | 265.500000 | 268.989990 | 258.706665 | 260.510010 | 260.510010 | 91449900 | TSLA |
| 3 | 2021-10-05 | 261.600006 | 265.769989 | 258.066681 | 260.196655 | 260.196655 | 55297800 | TSLA |
| 4 | 2021-10-06 | 258.733337 | 262.220001 | 257.739990 | 260.916656 | 260.916656 | 43898400 | TSLA |
| ... | ... | ... | ... | ... | ... | ... | ... | ... |
| 6295 | 2022-09-23 | 13.090000 | 13.892000 | 12.860000 | 13.710000 | 13.710000 | 28279600 | XPEV |
| 6296 | 2022-09-26 | 14.280000 | 14.830000 | 14.070000 | 14.370000 | 14.370000 | 27891300 | XPEV |
| 6297 | 2022-09-27 | 14.580000 | 14.800000 | 13.580000 | 13.710000 | 13.710000 | 21160800 | XPEV |
| 6298 | 2022-09-28 | 13.050000 | 13.421000 | 12.690000 | 13.330000 | 13.330000 | 31799400 | XPEV |
| 6299 | 2022-09-29 | 12.550000 | 12.850000 | 11.850000 | 12.110000 | 12.110000 | 33044800 | XPEV |

**Figure 1.** Stock dataset.

3.1.2. Text Dataset

The text data used in this study were tweets collected from Twitter (currently known as X) by Yukhymenko [11], which contained four attributes (namely, date, tweet, stock name, and company name, as shown in Figure 2). The dataset contained 80,792 observations and, for the purpose of this study, only tweets with more than 10 likes were selected.

**Figure 2.** Text dataset.

*3.2. Our Approach*

This paper deployed GRU<sub>vader</sub> as a hybrid approach for stock price prediction. GRU<sub>vader</sub> combines the power of a deep learning algorithm and lexicon-based sentiment analysis for prediction. This paper explains the two components below.

A gated recurrent unit (GRU) is a type of RNN designed as a simple alternative to LSTM networks. A GRU utilizes a gating mechanism that enables it to selectively update its hidden state at each time step, effectively learning both short- and long-term dependencies in sequential data. This makes a GRU effective for applications such as language translation, speech recognition, and time series forecasting. A GRU comprises two gates (namely, a reset gate and an update gate) to control the flow of information. The reset gate is used to determine how much of the previous hidden state should be forgotten. The update gate is used to control how much new information should be added to the hidden state. These gates allow GRUs to selectively retain or discard information, enabling the effective learning of temporal patterns while using fewer parameters than LSTMs, resulting in a simpler and computationally efficient model. Thus, a GRU can be mathematically represented as follows:

$$\text{Reset Gate } (r_t) = (W_r \cdot [h_{t-1}, x_t] + b_r) \quad (1)$$

$$\text{Update Gate } (u_t) = \sigma(W_u \cdot [h_{t-1}, x_t] + b_u) \quad (2)$$

$$\text{Candidate Activations } (h_t) = \tanh(W_h \cdot [r_t * h_{t-1}, x_t] + b_h) \quad (3)$$

$$\text{Output } (h_t) = (1 - u_t) * h_{t-1} + u_t * \hat{h}_t \quad (4)$$

where $x_t$ is the input vectors at time t; $h_t$ is the new hidden state at time t; $h_{t-1}$ is the previous hidden state; $W_r$, $W_u$, and $W_h$ are the weight matrices for the reset gate, update gate, and candidate activation, respectively; $b_r$, $b_u$, and $b_h$ are the bias vectors for the reset gate, update gate, and candidate activation, respectively; $\sigma$ is the sigmoid activation function; and Tanh is the hyperbolic tangent activation function.

3.2.1. Sentiment Analysis

In this context, a sentiment analysis is the process of classifying public or investor emotions and opinions expressed in text into categories such as positive, negative, and neutral. There are two popular approaches to a sentiment analysis, namely, an ML approach and a lexicon-based approach [49–51]. This study utilized the lexicon-based approach because a supervised ML approach requires a large, labelled dataset for training purposes. Unfortunately, this is not readily available. Based on this, we employed the VADER lexicon to analyze the sentiment of the text data. This helped to measure the exogenous features related to stock. VADER (Valence Aware Dictionary and Sentiment Reasoner) was specifically designed for social media content and uses a lexicon and rule-based approach to analyze the polarity and intensity of emotions. The VADER lexicon comprises 7517 words and phrases, each of which are assigned sentiment scores ranging from −4 to +4. The VADER algorithm breaks down text into tokens and, based on the labels in the lexicon, it assigns an overall sentiment score (normalized compound score) in a range of −1 to +1.

3.2.2. Experimental Setup

The GRU was implemented in Python using the Keras library (https://colab.research.google.com/, accessed on 9 October 2024). In the GRU architecture, we used two GRU layers. The first layer was a GRU with 50 units and the second layer comprised a dropout layer (20%) to prevent overfitting and an output layer to predict the adjusted closing price. Further to this, we performed hyperparameter tuning on the number of units, dropout rate, and learning rate using a random search approach to minimize validation loss. The optimal model configuration was selected based on the performance with the validation set and training was enhanced using callbacks like EarlyStopping and ReduceLROnPlateau to effectively improve and avoid overfitting. The Adam Optimizer was chosen for training due to its adaptive learning rates and computational efficiency. Adam combines the features of Adagrad and RmSProp, making it suitable for the optimization of non-stationary objectives with sparse gradients, which helps to achieve faster convergences and more stable parameter updates compared with traditional optimizers. Thus, we could summarize our experimentation as follows, and the workflow is presented in Figure 3.

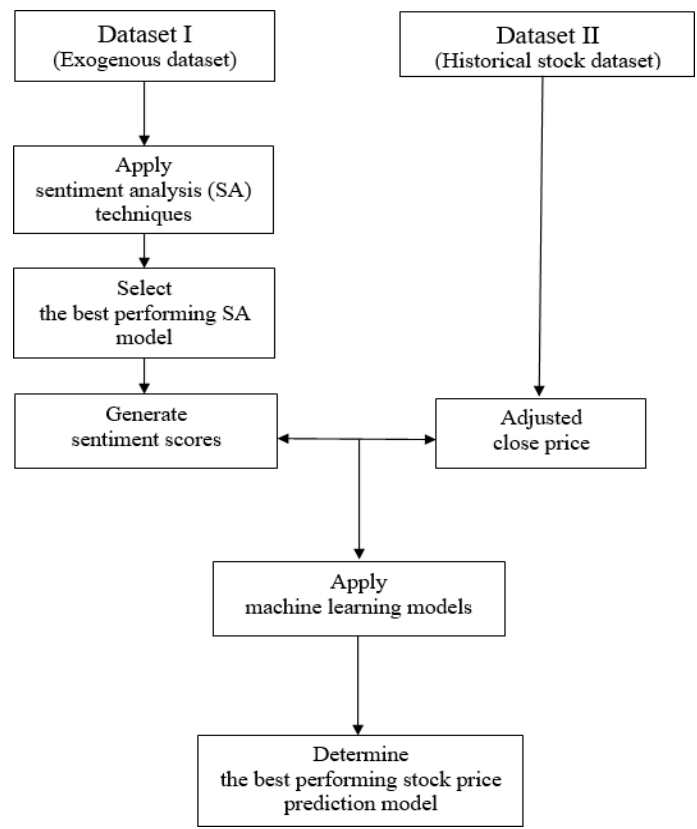

**Figure 3.** An illustration of our experiment.

- Firstly, we experimented with a GRU without applying a sentiment indicator and compared the results with other ML algorithms (LSTM, Bi-LSTM, Bi-GRU, and GAN), including the traditional time series model ARIMA.
- The tweets were pre-processed and we performed the sentiment analysis using VADER. Furthermore, we compared the VADER results with the AFINN and TextBlob results to determine the best-performing lexicon for this purpose.
- Moreover, we calculated the correlation between the adjusted close price and sentiment score to identify which stock exhibited the best correlation using the sentiment score and the price movement. This measured the behavior of exogenous features towards the adjusted close price.
- Lastly, we experimented with a hybrid of the GRU and the sentiment indicator, and compared the results with the other ML algorithms (LSTM, Bi-LSTM, Bi-GRU, and GAN) integrated with the sentiment scores, including the traditional time series model ARIMA.

### 3.3. Evaluation Metrics I

For the evaluation of the sentiment analysis lexicons, we randomly selected 2000 tweets (1225 positive, 263 neutral, and 511 negative) and manually labelled them as positive, negative, or neutral to create a reference dataset. In cases of disagreement, the tweets were discussed within the team to agree on the label. Thus, the 2000 labelled tweets were used to validate the performance of the lexicons. A confusion matrix was used to evaluate the performance of the lexicons by comparing the actual values with the predicted results. This was useful to calculate the statistical metrics, namely, true-positive (TP), true-negative (TN), false-positive (FP), and false-negative (FN). These metrics could be described as follows:

- True-positive (TP): The number of instances where the model correctly predicted a positive outcome.
- False-positive (FP): The number of instances where the model incorrectly predicted a positive outcome when it should have been negative.
- True-negative (TN): The number of instances where the model correctly predicted a negative outcome.
- False-negative (FN): The number of instances where the model incorrectly predicted a negative outcome when it should have been positive.

The statistical metrics were used to calculate the classification metrics (namely, accuracy, precision, recall, and F1-score), which could be described as follows:

Accuracy: Measurement of how often the model correctly predicted the outcome.

$$\text{Accuracy} = \frac{TP + TN}{TP + TN + FP + FN} \quad (5)$$

Recall: Measurement of the actual positive instances that were correctly predicted by the model.

$$\text{Recall} = \frac{TP}{TP + FN} \quad (6)$$

Precision: Measurement of the positive predictions that were actually correct.

$$\text{Precision} = \frac{TP}{TP + FP} \quad (7)$$

F1-score: The harmonic means of precision and recall, providing a balance between the two metrics.

$$F1 - \text{score} = 2 \times \frac{Precision \times recall}{Precision + recall} \quad (8)$$

### 3.4. Evaluation Metrics II

For the evaluation of the regression models, we employed the R-squared, mean square error, and the mean absolute error. These metrics could be described as follows:

Mean absolute error (MAE): The MAE measured the average of the absolute differences between the predicted and actual values, providing an intuitive reflection of prediction errors.

$$\text{MAE} = \frac{1}{n}\sum_{i=1}^{n} |y_i - \hat{y}_i| \tag{9}$$

Mean square error (MSE): The MSE measured the average squared difference between the predicted and actual values, penalizing large errors more than MAE.

$$\text{MSE} = \frac{1}{n}\sum_{i=1}^{n} (y_i - \hat{y}_i)^2 \tag{10}$$

Adjusted R-squared: The adjusted $R^2$ values were based on the number of predictors, ensuring it accounted for the model complexity.

$$\text{Adjusted} - R^2 = 1 - \left[\frac{1 - R^2 \ (n-1)}{n - k - 1}\right] \tag{11}$$

## 4. Results

This section presents our experimental results in a two-fold manner. First, we present the results of the lexicon-based sentiment analysis. Second, we present the stock market prediction results.

### 4.1. Sentiment Analysis Results

To choose the best NLP model for the sentiment analysis of tweets, 2000 tweets were randomly selected and manually labelled as positive, negative, or neutral. These manually labelled sentiments were than compared with the predictions from three models (AFINN, TextBlob, and VADER). The models were evaluated based on accuracy, precision, recall, and F1-score, as shown in Table 1.

**Table 1.** Evaluation results I: sentiment analysis.

| Lexicon  | Accuracy | Precision | Recall | F1-Score |
|----------|----------|-----------|--------|----------|
| AFINN    | 49.1     | 59.8      | 49.1   | 52.3     |
| TextBlob | 46.9     | 57.1      | 46.9   | 49.7     |
| VADER    | 54.1     | 61.0      | 54.1   | 56.4     |

### 4.2. Correlation

Figure 4 shows the correlation coefficients of various stocks. The results indicated that META and TSLA had the highest correlation coefficients, which were moderate. The order was as follows: TSLA (0.44), META (0.44), BA (0.33), PG (0.21), NIO (0.29), AMD (0.30), AAPL (0.15), and PYPL (0.28). The correlation coefficients showed that, in general, the sentiment indicator provided a moderate to low impact on the price movement. However, it is worth stating that the analysis was based on the sentiment scores of daily tweets. Aggregating scores on a weekly or monthly basis may have yielded a stronger correlation. The volume of tweets (as shown in Table 2) may also have affected the correlation coefficients. We selected the sentiment indicator variable only to assess if the stock price had any impact on public sentiment.

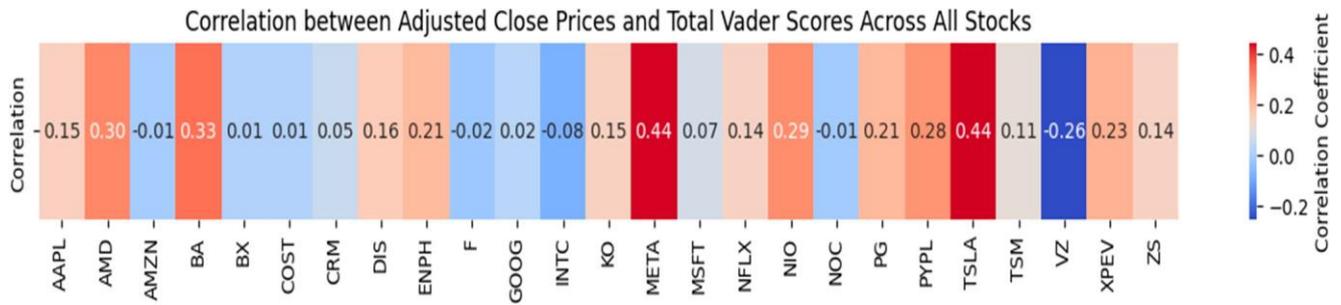

**Figure 4.** Correlation coefficient of stocks.

**Table 2.** Tweets and correlation coefficients.

| Stock Name | Volume of Tweets | Correlation Coefficient |
| --- | --- | --- |
| TSLA | 37422 | 0.44 |
| AAPL | 5056 | 0.15 |
| BA | 399 | 0.33 |
| META | 2751 | 0.44 |
| NIO | 3021 | 0.29 |
| PG | 4089 | 0.21 |
| AMD | 2227 | 0.30 |

Table 2 indicates that the sentiment scores did not only depend on the volume of the tweets, but also on the strength of the sentiment. For instance, even with a small volume of tweets (399), BA had a moderate correlation coefficient.

Figure 5 compares the risk and expected return of different stocks without taking sentiment into account. In this scenario, TSLA offered high returns but carried a high risk, indicating significant volatility. On the other hand, stocks like PG and AAPL showed a lower risk and lower expected returns.

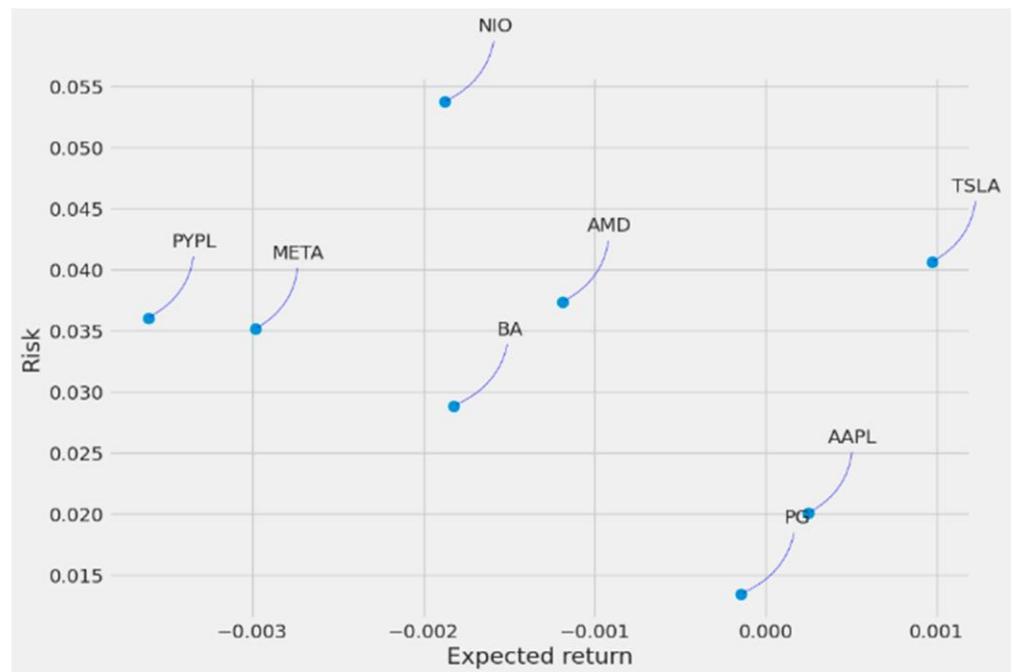

**Figure 5.** Risk vs. expected return without sentiment indicator.

Figure 6 examines the effect of sentiment on risk and expected return. The correlation analysis revealed that TSLA and META had moderate correlations with sentiment scores, while AMD, BA, NIO, PG, and PYPL showed weaker correlations. Figure 6 indicates that the TSLA, AMD, and NIO returns increased when sentiment was considered, and a change in risk was observed across other stocks as well. PG showed an increase in risk, which highlights the impact of sentiment on stock prediction. Thus, we needed to consider TSLA, AMD, META, NIO, and PG stocks, which exhibited a positive relationship with sentiment. In our analysis, using ML models helped us to better understand the impact of sentimental scores.

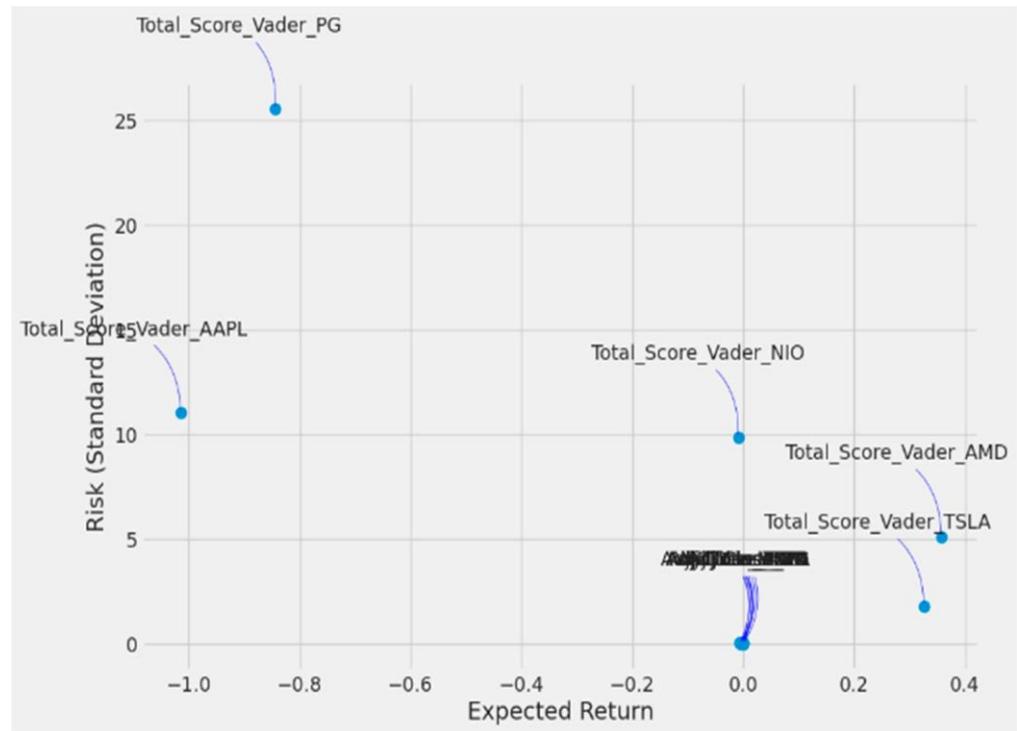

**Figure 6.** Risk vs. expected return with sentiment indicator.

*4.3. Model Performance*

This section presents the results of our experiment on stock prediction using various ML models, both without and with the incorporation of a sentimental analysis. We evaluated the performance of the models using metrics such as adjusted $R^2$, MAE, MSE, and accuracy. Tables 3 and 4 summarize the evaluation results.

**Table 3.** Evaluation results II: stock price prediction.

| Stock Name | Model | Adjusted $R^2$ | MAE | MSE | Accuracy |
|---|---|---|---|---|---|
| TSLA (Tesla) | ARIMA | −2.53 | 40.45 | 1798.23 | 86.22% |
| | LSTM (1_Layer) | −0.10 | 10.06 | 140.37 | 96.54% |
| | LSTM (2_Layer) | −0.18 | 10.53 | 151.06 | 96.39% |
| | Bi-LSTM (1_Layer) | 0.01 | 9.54 | 125.88 | 96.72% |
| | LSTM (Hyperparameter Tuning) | 0.28 | 7.82 | 91.45 | 97.29% |
| | GRU (1_Layer) | 0.10 | 4.02 | 114.80 | 96.87% |
| | GRU (2_Layer) | −0.10 | 0.10 | 140.69 | 96.54% |
| | Bi-GRU (1_Layer) | −0.03 | 9.65 | 131.51 | 96.66% |

**Table 3.** *Cont.*

| Stock Name | Model | Adjusted R² | MAE | MSE | Accuracy |
|---|---|---|---|---|---|
| TSLA (Tesla) | GRU (Hyperparameter Tuning) | 0.39 | 7.12 | 77.57 | 97.53% |
| | GAN (LSTM + CNN) | 0.53 | 6.72 | 78.98 | 97.65% |
| | GAN (GRU + CNN) | 0.46 | 7.24 | 89.20 | 97.47% |
| NIO | ARIMA | −0.85 | 1.40 | 3.50 | 93.03% |
| | LSTM (1_Layer) | −1.01 | 1.69 | 4.13 | 90.85% |
| | LSTM (2_Layer) | −0.12 | 1.12 | 2.29 | 93.55% |
| | Bi-LSTM (1_Layer) | 0.15 | 1.06 | 1.75 | 94.50% |
| | LSTM (Hyperparameter Tuning) | 0.41 | 0.85 | 1.22 | 95.60% |
| | GRU (1_Layer) | −0.95 | 1.65 | 3.99 | 91.05% |
| | GRU (2_Layer) | 0.20 | 1.02 | 1.65 | 94.69% |
| | Bi-GRU (1_Layer) | −1.30 | 1.92 | 4.72 | 89.78% |
| | GRU (Hyperparameter Tuning) | 0.39 | 0.85 | 1.24 | 95.54% |
| | GAN (LSTM + CNN) | 0.62 | 0.66 | 0.72 | 96.63% |
| | GAN (GRU + CNN) | 0.58 | 0.70 | 0.79 | 96.40% |
| AMD | ARIMA | 0.25 | 8.05 | 89.21 | 90.55% |
| | LSTM (1_Layer) | 0.37 | 7.37 | 82.59 | 90.54% |
| | LSTM (2_Layer) | 0.76 | 4.98 | 31.93 | 93.95% |
| | Bi-LSTM (1_Layer) | 0.50 | 6.62 | 65.87 | 91.53% |
| | LSTM (Hyperparameter Tuning) | 0.88 | 3.55 | 16.26 | 95.78% |
| | GRU (1_Layer) | 0.29 | 7.83 | 92.77 | 89.97% |
| | GRU (2_Layer) | 0.84 | 4.07 | 20.94 | 95.11% |
| | Bi-GRU (1_Layer) | 0.16 | 9.13 | 110.78 | 88.56% |
| | GRU (Hyperparameter Tuning) | 0.90 | 3.17 | 13.53 | 96.23% |
| | GAN (LSTM + CNN) | 0.91 | 2.65 | 10.77 | 97.91% |
| | GAN (GRU + CNN) | 0.91 | 2.60 | 10.20 | 96.96% |
| META | ARIMA | 0.49 | 6.83 | 76.34 | 95.79% |
| | LSTM (1_Layer) | 0.14 | 9.25 | 129.71 | 93.92% |
| | LSTM (2_Layer) | 0.21 | 9.23 | 119.51 | 94.25% |
| | Bi-LSTM (1_Layer) | 0.33 | 8.45 | 101.58 | 94.72% |
| | LSTM (Hyperparameter Tuning) | 0.78 | 4.89 | 32.82 | 96.92% |
| | GRU (1_Layer) | 0.16 | 9.17 | 127.40 | 94.18% |
| | GRU (2_Layer) | 0.76 | 5.22 | 36.93 | 96.73% |
| | Bi-GRU (1_Layer) | −3.51 | 25.21 | 682.93 | 83.88% |
| | GRU (Hyperparameter Tuning) | 0.80 | 4.54 | 29.58 | 97.14% |
| | GAN (LSTM + CNN) | 0.78 | 4.60 | 33.45 | 97.08% |
| | GAN (GRU + CNN) | 0.78 | 4.60 | 33.16 | 97.09% |

**Table 4.** Evaluation results III: stock price prediction (with sentiment indicator).

| Stock Name | Model | Adjusted R² | MAE | MSE | Accuracy |
|---|---|---|---|---|---|
| TSLA (Tesla) | ARIMA + Vader | −7.19 | 34.53 | 1357.36 | 88.00% |
|  | LSTM (1_Layer) + Vader | 0.06 | 9.49 | 117.01 | 96.74% |
|  | LSTM (2_Layer) + Vader | −0.09 | 10.03 | 136.37 | 96.54% |
|  | Bi-LSTM (1_Layer) + Vader | 0.06 | 9.39 | 116.37 | 96.79% |
|  | LSTM (Hyperparameter Tuning) + Vader | 0.02 | 9.40 | 121.28 | 96.76% |
|  | GRU (1_Layer) + Vader | 0.11 | 9.50 | 110.88 | 96.74% |
|  | GRU (2_Layer) + Vader | 0.09 | 9.08 | 113.00 | 96.87% |
|  | Bi-GRU (1_Layer) + Vader | 0.02 | 9.40 | 121.78 | 96.77% |
|  | **GRU$_{vader}$** | 0.40 | 6.93 | 74.23 | 97.60% |
|  | GAN (LSTM + CNN) + Vader | −1.33 | 15.83 | 385.86 | 94.48% |
|  | GAN (GRU + CNN) + Vader | −0.55 | 13.41 | 256.73 | 95.42% |
| NIO | ARIMA + Vader | −6.12 | 3.06 | 13.17 | 84.37% |
|  | LSTM (1_Layer) + Vader | 0.10 | 1.13 | 2.22 | 93.92% |
|  | LSTM (2_Layer) + Vader | −0.10 | 1.21 | 2.22 | 93.66% |
|  | Bi-LSTM (1_Layer) + Vader | 0.09 | 1.09 | 1.81 | 94.36% |
|  | LSTM (Hyperparameter Tuning) + Vader | 0.16 | 0.96 | 1.67 | 95.09% |
|  | GRU (1_Layer) + Vader | 0.15 | 0.95 | 1.69 | 95.03% |
|  | GRU (2_Layer) + Vader | 0.07 | 1.07 | 1.86 | 94.37% |
|  | Bi-GRU (1_Layer) + Vader | 0.01 | 1.14 | 1.97 | 93.99 |
|  | **GRU$_{vader}$** | 0.50 | 0.73 | 0.99 | 96.14% |
|  | GAN (LSTM + CNN) + Vader | −10.06 | 4.30 | 20.88 | 78.29% |
|  | GAN (GRU + CNN) + Vader | 0.20 | 0.98 | 1.51 | 94.87% |
| AMD | ARIMA + Vader | −0.01 | 9.87 | 118.12 | 88.88% |
|  | LSTM (1_Layer) + Vader | 0.39 | 7.35 | 77.01 | 90.95% |
|  | LSTM (2_Layer) + Vader | 0.73 | 5.18 | 34.42 | 93.70% |
|  | Bi-LSTM (1_Layer) + Vader | 0.84 | 3.96 | 19.90 | 95.21% |
|  | LSTM (Hyperparameter Tuning) + Vader | 0.86 | 3.57 | 17.50 | 95.73% |
|  | GRU (1_Layer) + Vader | 0.17 | 8.24 | 105.93 | 89.65% |
|  | GRU (2_Layer) + Vader | 0.43 | 7.15 | 72.62 | 96.95% |
|  | Bi-GRU (1_Layer) + Vader | −0.05 | 10.36 | 135.62 | 87.04% |
|  | **GRU$_{vader}$** | 0.91 | 2.70 | 10.73 | 96.86% |
|  | GAN (LSTM + CNN) + Vader | 0.78 | 4.47 | 26.47 | 94.63% |
|  | GAN (GRU + CNN) + Vader | 0.14 | 8.70 | 102.67 | 89.11% |

**Table 4.** *Cont.*

| Stock Name | Model | Adjusted R² | MAE | MSE | Accuracy |
|---|---|---|---|---|---|
| META | ARIMA + Vader | 0.40 | 7.84 | 87.59 | 95.02% |
| | LSTM (1_Layer) + Vader | 0.02 | 10.21 | 144.49 | 93.64% |
| | LSTM (2_Layer) + Vader | 0.23 | 8.62 | 112.78 | 94.55% |
| | Bi-LSTM (1_Layer) + Vader | 0.31 | 8.20 | 100.87 | 94.70% |
| | LSTM (Hyperparameter Tuning) + Vader | 0.77 | 4.91 | 33.12 | 96.91% |
| | GRU (1_Layer) + Vader | −3.05 | 21.96 | 599.92 | 85.67% |
| | GRU (2_Layer) + Vader | 0.71 | 5.62 | 42.47 | 96.45% |
| | Bi-GRU (1_Layer) + Vader | 0.69 | 5.74 | 44.72 | 96.38% |
| | **GRU$_{vader}$** | 0.81 | 4.30 | 26.69 | 97.30% |
| | GAN (LSTM + CNN) + Vader | −2.62 | 21.70 | 544.10 | 86.07% |
| | GAN (GRU + CNN) + Vader | −8.47 | 37.07 | 1421.90 | 77.14% |

The initial analysis (results shown in Table 3) focused on predicting stock prices using a historical stock dataset without incorporating a sentiment indicator. Across all the stocks, the traditional ARIMA model underperformed compared with the neural network models. For example, with TSLA, ARIMA achieved an adjusted R² of −2.5265 and an MAE of 40.45, which represented the poorest achieved performance.

The neural network models achieved superior performance. For example, the LSTM model with hyperparameter tuning achieved an adjusted R² of 0.2829 and an MAE of 7.826. The GRU model with hyperparameter tuning produced improved results, with an adjusted R² of 0.39294 and an MAE of 7.122. The GAN (LSTM + CNN) achieved an adjusted R² of 0.53 and an MAE of 6.72. This performance result trend was observed across other stocks. In summary, our first-stage experiment showed that the GAN models and the GRU with hyperparameter tuning achieved the best results, which were similar and consistent across all the metrics and stocks.

In the second experiment (results shown in Table 4), we incorporated the sentiment scores as continuous values rather than categoricals. As such, this formed a new merged dataset, which became the final merged dataset used to fit the models. For all the stocks, GRU$_{vader}$ showed the best performance. For example, TSLA with an adjusted R² increased from 0.39294 to 0.40526 and the MAE decreased from 7.122 to 6.939. Improvements were generally observed across the stocks, which also yielded a positive impact on the model's predictive performance. However, it is worth stating that the GAN models in this second experiment showed poor performance. The adjusted R² dropped to negative values and the MAE and MSE significantly increased, suggesting that GAN models may not effectively predict trends with sentiment features.

In the NIO stock, the inclusion of a sentiment indicator led to notable improvements. GRU$_{vader}$ showed an adjusted R² of 0.50236 (up from 0.3959 without a sentiment) and the MAE decreased from 0.853 to 0.739. The accuracy improved to 96.14%, highlighting the beneficial effect of a sentiment analysis on predicting the NIO stock price.

For AMD, GRU$_{vader}$ showed enhanced performance with a sentiment analysis. The adjusted R² increased from 0.89676 to 0.91619 and the MAE reduced from 3.146 to 2.705.

For META, GRU$_{vader}$ also performed well. The adjusted R² improved from 0.80416 to 0.81954 and the MAE decreased from 4.546 to 4.307. The accuracy increased to 97.30%, suggesting that incorporating the sentiment score enhanced the model's ability to predict stock price.

The results showed that incorporating a sentiment analysis could enhance the predictive performance of certain models for stock prediction. The GRU$_{vader}$ model consistently outperformed other models across multiple stocks. The improvement in the adjusted $R^2$ and reduced MSE and MAE values indicated that the sentiment analysis provided additional valuable information, enabling the models to capture market dynamics more effectively. This was particularly evident for stocks like TSLA, META, NIO, and AMD where public sentiment appeared to have a strong correlation with stock price movement. In addition, we compared our results with existing results in the literature and we observed (as shown in Table 5) that GRU$_{vader}$ yielded improved results.

**Table 5.** Comparison with previous results.

| Paper | Model | MSE | RMSE | MAE | Adj-$R^2$ | Accuracy | Stock Name |
|---|---|---|---|---|---|---|---|
| 9 | Bi-LSTM | - | - | 0.07121 | - | - | - |
| 10 | Bi-LSTM | 0.0355 | 0.188206 | - | - | 94% | - |
| 11 | GAN | - | - | - | - | - | - |
| Our Study | GRU$_{vader}$ | 0.124 | 0.0498 | 0.00248 | 0.40526 | 97.60% | TSLA |
|  |  | 0.34 | 0.0583 | 0.0034 | 0.91619 | 96.86% | AMD |
|  |  | 0.24 | 0.0490 | 0.0024 | 0.81954 | 97.30% | META |
|  |  | 0.10 | 0.0316 | 0.0010 | 0.50236 | 96.14% | NIO |

Furthermore, we produced residual plots to show the residuals (errors) of the GRU$_{vader}$ predictions compared with the actual values. The plots shown in Figures 7–10 are residual plots generated for each stock (TSLA, META, NIO, and AMD) after applying hyperparameter tuning with the sentiment analysis to investigate the GRU$_{vader}$ performances. From Figures 7–10, we assessed the performance of our GRU$_{vader}$ using the following indicators:

- Q–Q (Quantile–quantile plots): The residuals should follow a normal distribution, and thus the residuals should closely align with the reference line in the Q–Q plot.
- Observation plot: To check the randomness of the residuals, residuals should randomly scatter without following any patterns. The randomness indicates that the model has effectively captured the trends.

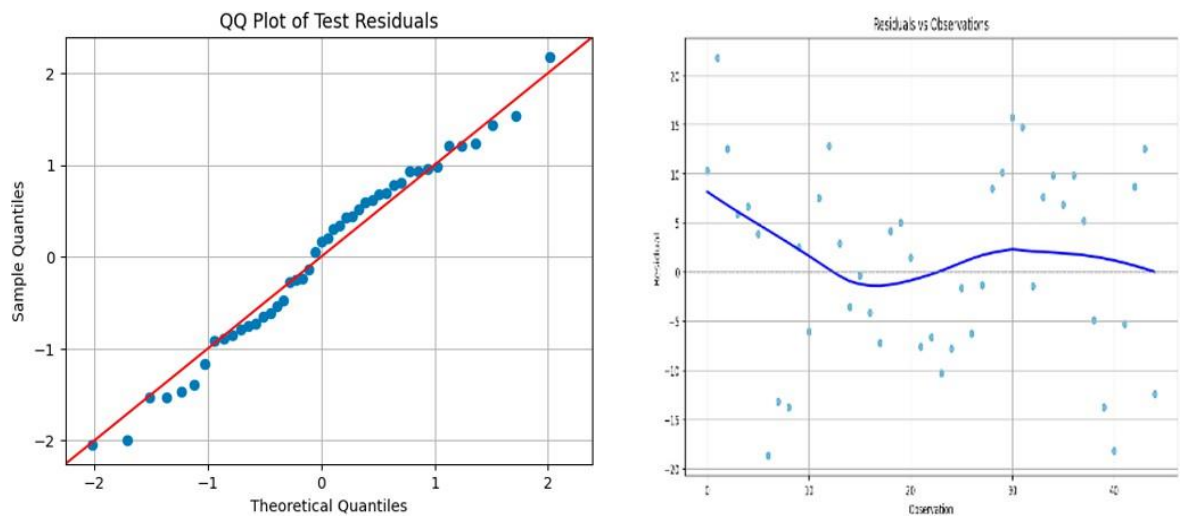

**Figure 7.** Residual plots for stocks: TSLA.

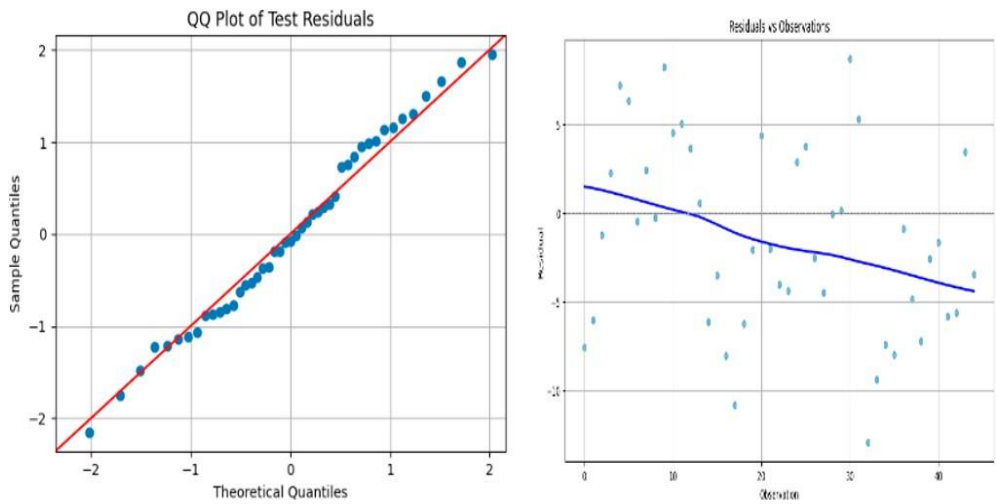

**Figure 8.** Residual plots for stocks: META.

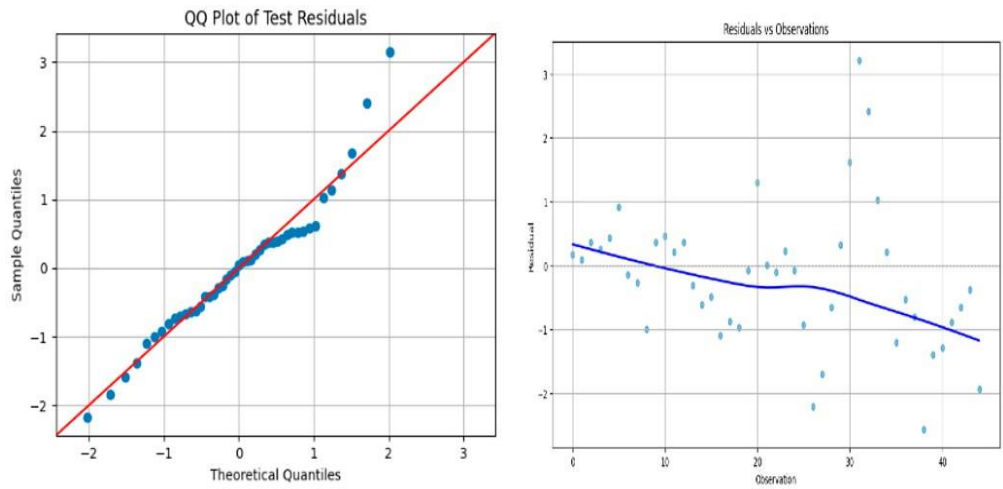

**Figure 9.** Residual plots for stocks: NIO.

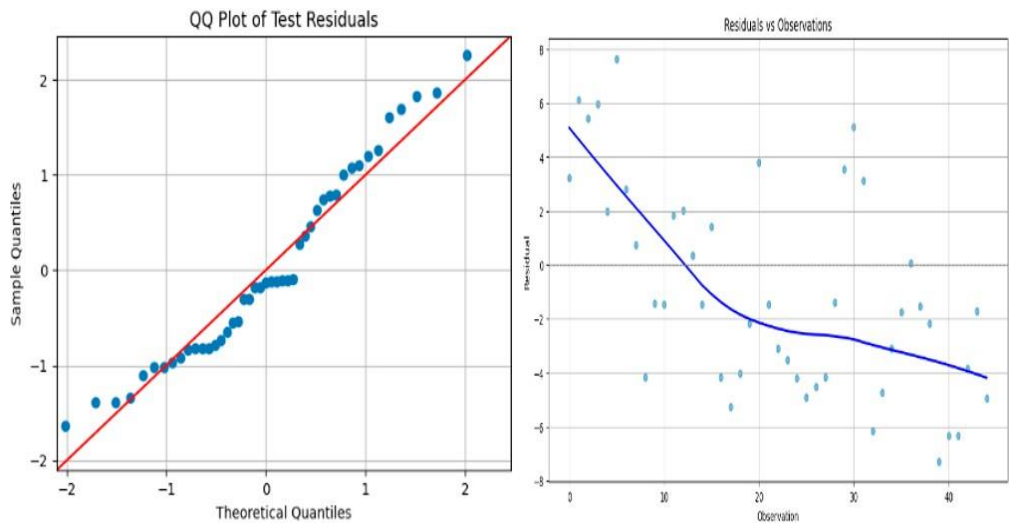

**Figure 10.** Residual plots for stocks: AMD.

## 5. Conclusions

This paper aimed to develop a stock price prediction model with improved performance. In addition, we examined the influence of exogenous features in the form of public sentiment. In doing so, several ML algorithms, including ARIMAX, LSTM, GRU, Bi-LSTM, Bi-GRU, and GAN, were used to predict stock prices using both historic stock data and sentiment scores derived from tweets. Our results justified that incorporating a sentiment analysis significantly improves the performance of stock model prediction, particularly with GRU$_{vader}$. However, our experimental findings indicated that sentiment indicators did not improve performance for all algorithms, as was the case for the GAN in this study.

To conclude, this paper highlights the importance of exogenous features such as tweets in financial prediction and shows that sentiment indicators can play a crucial role in predicting stock price movement. Theoretically, this study provides evidence that a sentiment analysis can enhance stock prediction models. However, this is not the case for all algorithms. There is a need for continuous comparative studies in this domain due to the ongoing challenges in capturing complex market movement. Thus, this study identifies and outlines future work as follows:

- In future studies, we aim to develop a compound stock price prediction model that examines and incorporates more exogenous features (such as financial news, financial reports, other social media platforms, and geopolitical event indicators).
- Further to this, future studies should consider investigating the influence of senti- ments from other assets (for example, sentiments from cryptocurrency text) on stock price movement.
- Future work should consider integrating specific stock language models and deep learning algorithms to examine the ability of the hybrid approach to manage long-term dependencies and complex patterns.
- The study also emphasizes the need to extend this methodology to other sectors and explore its potential for identifying market trends during volatile conditions.

**Author Contributions:** Conceptualization: A.M. and B.O.; methodology: A.M., B.O., S.T.I. and O.S.; software: A.M.; validation: A.M. and B.O.; formal analysis: A.M.; investigation: A.M. and B.O.; resources: A.M. and B.O.; data curation: A.M. and B.O.; writing—original draft preparation: A.M. and B.O.; writing—review and editing: A.M., B.O. and S.T.I.; supervision: B.O.; project administration: B.O., S.T.I. and O.S. All authors have read and agreed to the published version of the manuscript.

**Funding:** This research received no external funding.

**Data Availability Statement:** The original contributions presented in the study are included in the article, further inquiries can be directed to the corresponding author.

**Conflicts of Interest:** The authors declare no conflicts of interest.